%% file: main.tex
\documentclass[10pt,twocolumn,letterpaper]{article}

\usepackage{cvpr}              


\definecolor{cvprblue}{rgb}{0.21,0.49,0.74}
\usepackage[pagebackref,breaklinks,colorlinks,allcolors=cvprblue]{hyperref}
\usepackage[accsupp]{axessibility} 

\usepackage{amsmath}
\input{math_commands.tex}

\def\paperID{2098} 
\def\confName{CVPR}
\def\confYear{2025}

\title{Zero-1-to-A: Zero-Shot One Image to Animatable Head Avatars Using Video Diffusion}

\author{
\textbf{Zhenglin Zhou}$^{1,2}$,
\textbf{Fan Ma}$^2$,
\textbf{Hehe Fan}$^{2,\dagger}$,
\textbf{Tat-Seng Chua}$^3$ \\
\textsuperscript{\rm 1} State Key Laboratory of Brain-machine Intelligence, Zhejiang University \\
\textsuperscript{\rm 2} ReLER, CCAI, Zhejiang University \\
\textsuperscript{\rm 3} National University of Singapore \\
{
\tt\small \{zhenglinzhou, mafan, hehefan\}@zju.edu.cn \quad dcscts@nus.edu.sg
}
}

\begin{document}
\maketitle
\def\thefootnote{$\dagger$}\footnotetext{Corresponding author.}


\input{sec/0_abstract}    
\input{sec/1_intro}

\input{sec/2_related_work}

\input{sec/3_preliminary}

\input{sec/3_method}

\input{sec/4_experiment}
\input{sec/5_conclusion}
\clearpage
\section*{Acknowledgements}
This work was supported in part by the National Natural Science Foundation of China (62472381), Fundamental Research Funds for the Zhejiang Provincial Universities (226-2024-00208), and the Natural Science Foundation of Zhejiang Province (LDT23F02023F02).

{
    \small
    \bibliographystyle{ieeenat_fullname}
    \bibliography{main}
}

\input{sec/X_suppl}

\end{document}

%% file: math_commands.tex
\usepackage{amsmath,amsfonts,bm}

\def\eqref#1{equation~\ref{#1}}

\def\1{\bm{1}}

\DeclareMathAlphabet{\mathsfit}{\encodingdefault}{\sfdefault}{m}{sl}
\SetMathAlphabet{\mathsfit}{bold}{\encodingdefault}{\sfdefault}{bx}{n}



%% file: sec/0_abstract.tex
\begin{abstract}

Animatable head avatar generation typically requires extensive data for training. 
To reduce the data requirements, a natural solution is to leverage existing data-free static avatar generation methods, such as pre-trained diffusion models with score distillation sampling (SDS), which align avatars with pseudo ground-truth outputs from the diffusion model. 
However, directly distilling 4D avatars from video diffusion often leads to over-smooth results due to spatial and temporal inconsistencies in the generated video.
To address this issue, we propose \textbf{Zero-1-to-A}, a robust method that synthesizes a spatial and temporal consistency dataset for 4D avatar reconstruction using the video diffusion model.
Specifically, Zero-1-to-A iteratively constructs video datasets and optimizes animatable avatars in a progressive manner, ensuring that avatar quality increases smoothly and consistently throughout the learning process. 
This progressive learning involves two stages: (1) Spatial Consistency Learning fixes expressions and learns from front-to-side views, and (2) Temporal Consistency Learning fixes views and learns from relaxed to exaggerated expressions, generating 4D avatars in a simple-to-complex manner. 
Extensive experiments demonstrate that Zero-1-to-A improves fidelity, animation quality, and rendering speed compared to existing diffusion-based methods, providing a solution for lifelike avatar creation.
Code is publicly available at: \href{https://github.com/ZhenglinZhou/Zero-1-to-A}{https://github.com/ZhenglinZhou/Zero-1-to-A}.

\end{abstract}

%% file: sec/1_intro.tex
\section{Introduction}
\label{sec:intro}

Head avatars are lifelike digital representations of humans created by computer graphics and deep learning. 
They can mimic human expressions, movements, and interactions, serving as speakers, assistants, characters, \etc in AR/VR, films and games.

Head avatar generation typically relies on large amounts of real or synthetic human data~\cite{wood2021fake, kirschstein2023nersemble, INSTA:CVPR2023, Zheng2023pointavatar, xu2023avatarmav, xie2022vfhq, zhu2022celebvhq, zhang2021flow}, which are not easy to collect.
Recently, score distillation-based methods~\cite{poole2022dreamfusion, wang2022sjc, hong2023debiasing, liu2023zero, liu2024one, long2024wonder3d, wang2023prolificdreamer, katzir2023noise, liang2024luciddreamer, wu2024consistent3d} have revolutionized this field by enabling the 3D/4D head avatar generation using pre-trained image diffusion models (\eg Stable Diffusion~\cite{stable_diffusion}) without the need of human data.
Existing methods mostly focus on text-conditioned generation, which is expressive but lacks the precision and control of visual inputs~\cite{bergman2023articulated, han2023headsculpt, liu2023headartist, zhou2024headstudio, wang2024headevolver}. 
Image inputs are critical for realistic 4D avatar generation but are still underexplored.

\begin{figure}[t]
    \centering
    \includegraphics[width=1.0\linewidth]{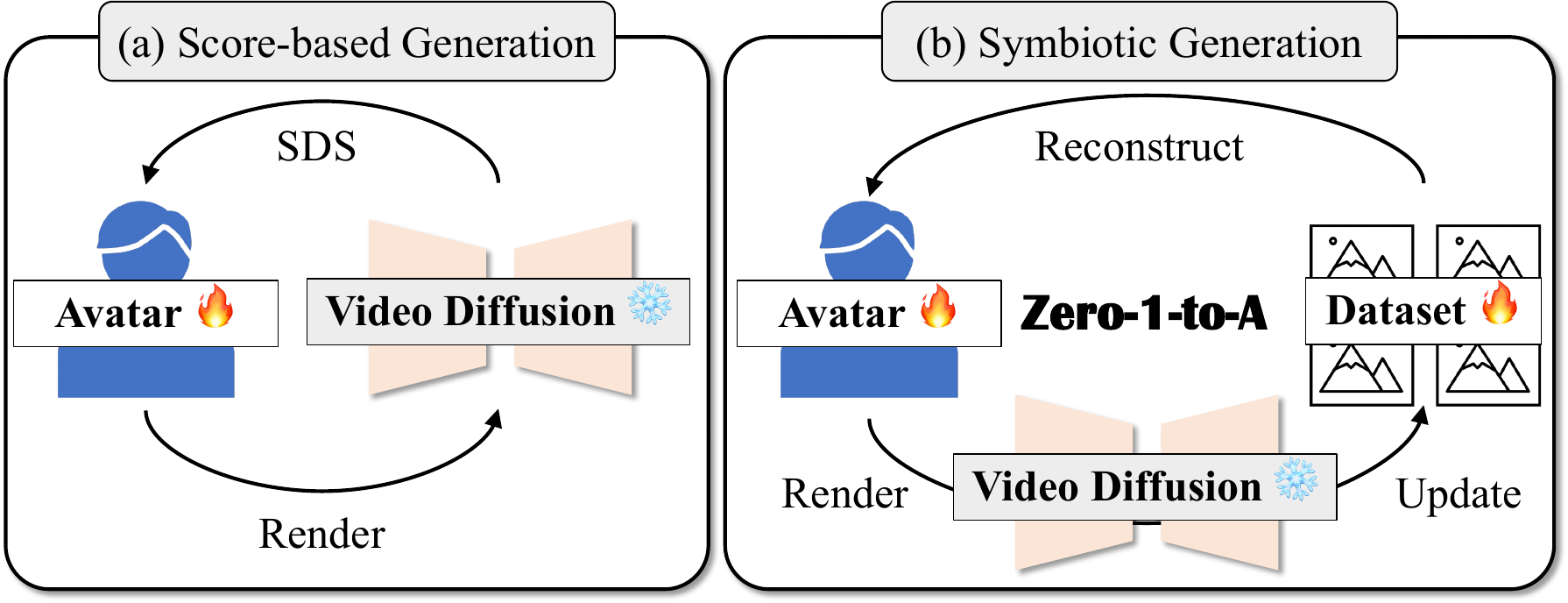}
    \caption{
    \textbf{4D Avatar generation with SDS loss~\cite{poole2022dreamfusion} and our Zero-1-to-A.}
    Video diffusion often suffers from spatial and temporal inconsistencies.
    (a) The SDS loss, aligning avatar with output from video diffusion, produces over-smooth results. 
    (b) Zero-1-to-A addresses this issue by synthesizing spatially and temporally consistent datasets for avatar reconstruction. 
    It introduces an updatable dataset to cache video diffusion results and establishes a mutually beneficial relationship between avatar generation and dataset construction to further enhance consistency.
    }
    \vspace{-1em}
    \label{fig:teaser}
\end{figure}

\begin{figure*}[t]
    \centering
    \includegraphics[width=1.0\linewidth]{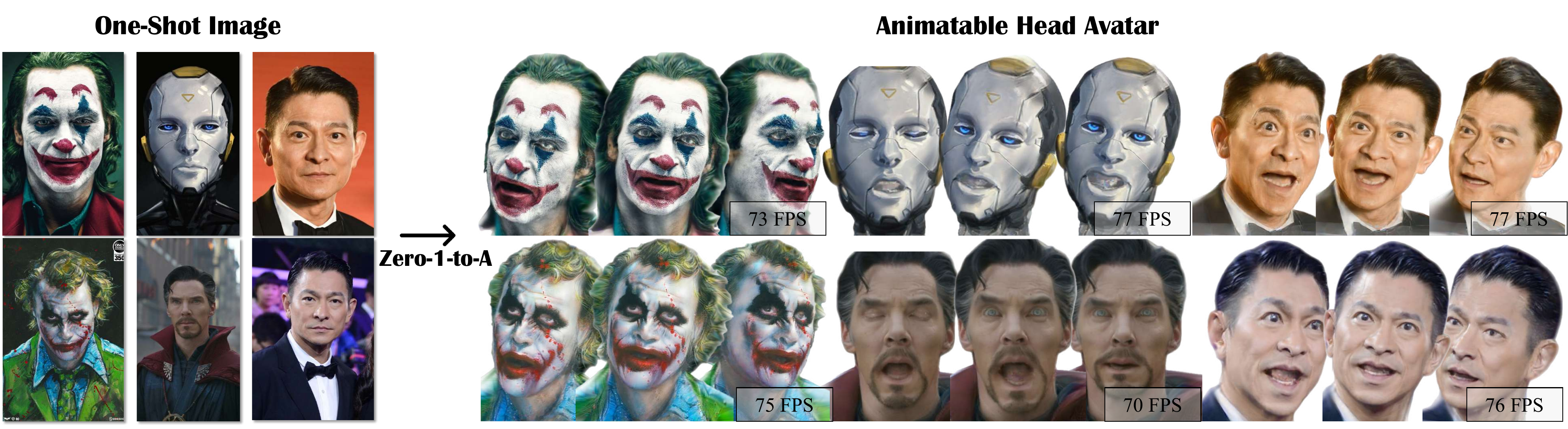}
    \vspace{-1.5em}
    \caption{
    \textbf{Image to animatable avatars generation by Zero-1-to-A}.
    Without manually annotated data, our method can distill high-fidelity 4D avatars with real-time rendering speed from a pre-trained video diffusion using only one image input.
    }
    \vspace{-0.5em}
    \label{fig:vis}
\end{figure*}

Recent advances in video diffusion~\cite{blattmann2023stable, wei2024aniportrait, chen2024echomimic, ma2024emoji} have made it feasible to generate dynamic head avatars by distilling from video diffusion models. 
However, as shown in \cref{fig:talking_head}, video diffusion suffers from spatial and temporal inconsistencies.
Essentially, SDS aims to align the images rendered by the 3D model with pseudo ground-truth outputs generated by the diffusion model~\cite{liang2024luciddreamer}.
Consequently, this instability in the distillation process often leads to over-smooth results, compromising the quality and fidelity of the avatars (refer to ``SDS Loss'' in \cref{fig:ablation}).

In this paper, we propose Zero-1-to-A, a novel approach for generating 4D avatars from a single image using a pre-trained video diffusion model.
Zero-1-to-A first introduces \textbf{SymGEN} which incorporates an updatable dataset to cache video diffusion results, reducing inconsistencies, and establishing a mutually beneficial cycle between avatar generation and dataset construction to enhance consistency.
Additionally, Zero-1-to-A employs a \textbf{Progressive Learning} strategy that advances from simple to complex scenarios.
This strategy decouples video diffusion generation into: 
(1) Spatial Consistency Learning, fixing expressions and learning from front-to-side views, and (2) Temporal Consistency Learning, fixing the camera and learning from relaxed to exaggerated expressions. 
Compared to classic random learning, progressive learning generates consistent results in simple views and expressions and thus better initializes the avatar.
As avatar quality improves, it further guides video diffusion to achieve greater consistency in more challenging cases, resulting in the smooth and continuous enhancement of avatar quality.

We evaluate Zero-1-to-A across various head avatar generation tasks, validating our design choices through comparisons with ablated variants of our method. 
Additionally, we qualitatively compare Zero-1-to-A to state-of-the-art head-specialized approaches~\cite{han2023headsculpt, liu2023headartist, zhou2024headstudio}. Comprehensive experiments show that Zero-1-to-A significantly enhances fidelity, animation quality, and rendering speed over existing diffusion-based methods, offering a robust solution for lifelike avatar creation. 

Overall, our contributions can be summarized as follows:
\begin{itemize}
    \item We present SymGEN, a novel method that gradually synthesizes spatial and temporal consistency datasets for 4D avatar reconstruction using the video diffusion model.
    \item We propose a Progressive Learning strategy, a decoupled learning strategy with Spatial and Temporal Consistency Learning that ensures stable initialization and smooth quality enhancement.
    \item Extensive experiments show that Zero-1-to-A achieves superior fidelity, animation quality, and rendering speed in 4D avatar generation compared with baseline methods.
\end{itemize}

%% file: sec/2_related_work.tex
\section{Related Work}
\label{sec:related_work}

\noindent\textbf{Object Generation.}
Recent advances in multi-modal alignment models~\cite{radford2021clip,yang2021multiple,luo2024mmevol,liu2024towards} and generative models~\cite{sohl2015deep, ho2020denoising,wang2024lavin} have led to remarkable progress in text-to-image generation~\cite{nichol2021glide, ho2022imagen, stable_diffusion}.
Building on this, Neural Radiance Fields (NeRF)~\cite{mildenhall2020nerf, barron2022mip} and 3D Gaussian Splatting~\cite{kerbl3Dgaussians} have opened new avenues for 3D-aware generation by enabling 3D scene reconstruction from only 2D multi-view images. 
Leveraging these techniques, recent methods integrate knowledge from text-to-image models to generate 3D content guided by text prompts~\cite{poole2022dreamfusion, wang2022sjc, hong2023debiasing, wang2023prolificdreamer, katzir2023noise, liang2024luciddreamer, wu2024consistent3d,zhou2025dreamdpo}.
Meanwhile, the image-conditioned generation has made significant progress~\cite{liu2023zero, liu2024one, long2024wonder3d}.
Moreover, the recent advancement of 3D generation also inspired multiple applications, including scenes generation~\cite{cohen2023set, hollein2023text2room}, 3D editing~\cite{Ayaan2023instructnerf, kamata2023instruct3d}, and avatar generation~\cite{cao2023dreamavatar,jiang2023avatarcraft,xu2023seeavatar,chen2023humanmac}.

\begin{figure*}[t]
    \centering
    \includegraphics[width=1.0\linewidth]{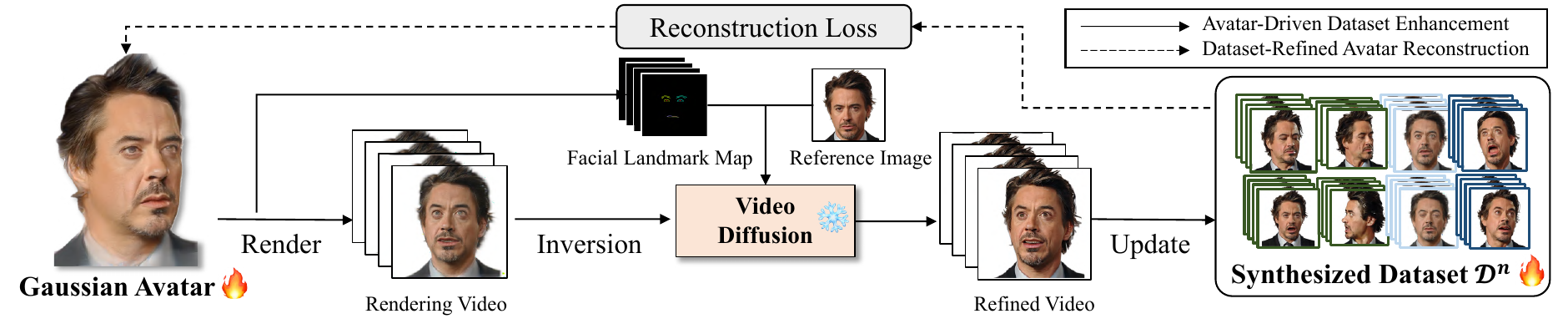}
    \vspace{-2.0em}
    \caption{
    \textbf{Framework of SymGEN.}
    Our method simultaneously builds both the dataset and avatar from scratch through video diffusion. 
    It establishes a mutually beneficial relationship between dataset construction and avatar reconstruction, iteratively updating the synthesized dataset and training the head avatar on the updated dataset to achieve unified results.
    }
    \label{fig:pipeline}
\end{figure*}

\noindent\textbf{Avatar Generation.}
Traditional 3D head avatar generation methods are based on statistical models such as 3DMM~\cite{Blanz19993dmm} and FLAME~\cite{li2017flame}, while recent approaches leverage 3D-aware Generative Adversarial Networks (GANs)\cite{schwarz2020graf, chan2021pigan, Chan2022eg3d, An2023panohead, zhang2023multi, Portrait3D_sig24}. 
Advances in dynamic scene representation~\cite{Gao-ICCV-DynNeRF, kplanes_2023, Cao2023HEXPLANE} have further improved animatable head avatar reconstruction. 
Given monocular or multi-view videos, methods like~\cite{zheng2022imavatar, INSTA:CVPR2023, Zheng2023pointavatar, xu2023avatarmav, qian2023gaussianavatars, kirschstein2023diffusionavatars,yan2024dialoguenerf} can reconstruct photo-realistic head avatars and animate them based on FLAME. 
However, these approaches rely heavily on large datasets of real or synthetic human data~\cite{wood2021fake, kirschstein2023nersemble, INSTA:CVPR2023, Zheng2023pointavatar, xu2023avatarmav, xie2022vfhq, zhu2022celebvhq, zhang2021flow,chu2024gagavatar,deng2024portrait4d}, which are challenging to collect.
Recently, diffusion-based methods have transformed this field by enabling 3D/4D head avatar generation using pre-trained image diffusion models without requiring human data.
Most of these methods focus on text-conditioned generation~\cite{bergman2023articulated, han2023headsculpt, liu2023headartist, zhou2024headstudio, wang2024headevolver}, leaving image-conditioned generation for realistic 4D avatar creation is largely underexplored.

%% file: sec/3_preliminary.tex
\section{Preliminaries}
\label{sec:pre}
In this section, we present a brief overview of the image-to-4D avatar generation. 
The generation process can be seen as distilling knowledge from a video diffusion model into a learnable 4D representation. 
Below, we briefly introduce the animatable Gaussian head and portrait video diffusion techniques used in our method.

\noindent\textbf{Animatable Gaussian Head}~\cite{qian2023gaussianavatars, zhou2024headstudio} is a 4D avatar representation that combines the animatable head model FLAME~\cite{li2017flame} with 3D Gaussian splatting (3DGS)~\cite{kerbl3Dgaussians} to enable high-quality texture and geometry modeling. 
Specifically, each 3D Gaussian point is rigged to the FLAME mesh, allowing the Gaussians to deform consistently with the mesh.
As a result, the 4D avatar can be effectively animated using FLAME's pose and expression parameters and efficiently rendered via the differentiable tile rasterizer~\cite{kerbl3Dgaussians}.

\noindent\textbf{Portrait Video Diffusion.}
Recent methods~\cite{wei2024aniportrait, chen2024echomimic, ma2024emoji} extend Stable Diffusion (SD)~\cite{stable_diffusion} for portrait video generation, creating videos where the identity matches the reference image and motion aligns with animation signals. 
Key modules include an appearance net for identity injection via self-attention in UNet, a motion injection module for motion control using ControlNet~\cite{zhang2023controlnet}, temporal attention with transformers for cross-frame consistency, and an image prompt injection module replacing CLIP’s text encoder with an image encoder to adapt SD for portrait animation.

\begin{figure*}[t]
    \centering
    \includegraphics[width=1.0\linewidth]{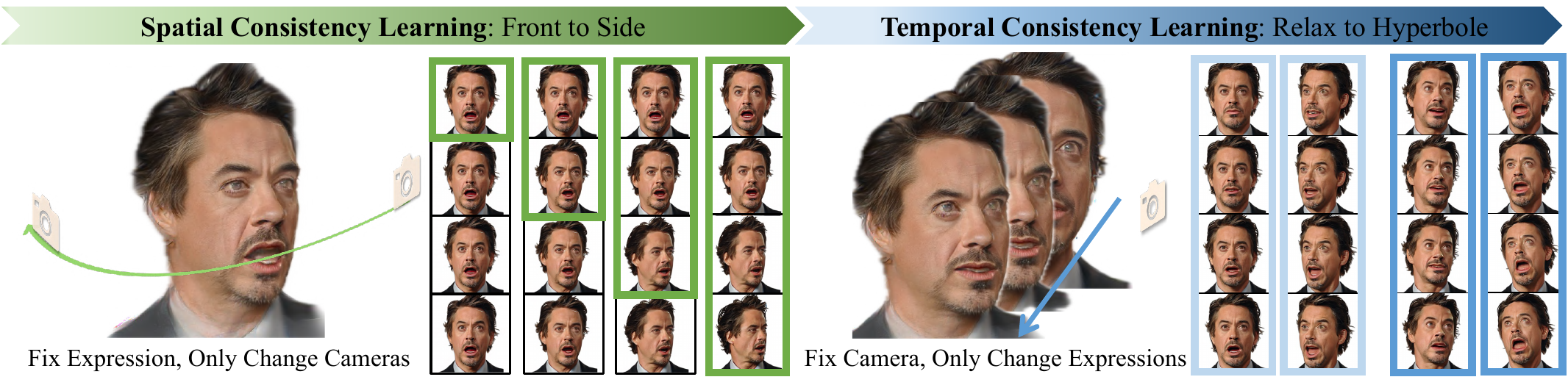}
    \caption{
    \textbf{Pipeline of Progressive Learning.}
    It sequences learning from simple to complex, facilitating symbiotic generation to create consistent avatars from inconsistent video diffusion. 
    This process divides 4D avatar generation into:
    (1) Spatial Consistency Learning: progressing from frontal to side views with a fixed expression.
    (2) Temporal Consistency Learning: learn from relaxed to hyperbole expressions under a fixed camera.
    }
    \vspace{-1.0em}
    \label{fig:progressive}
\end{figure*}

%% file: sec/3_method.tex
\section{Method}
\label{sec:method}

Zero-1-to-A is an image-to-4D avatar generation method.
Given a reference portrait, our goal is to generate an animatable head avatar using a pre-trained video diffusion.
The generation pipeline has two key components, including (1) the symbiotic generation in \cref{sec:symgen}, and (2) the progressive learning strategy in \cref{sec:prog-learn}.
Implementation details are discussed in \cref{sec:implementation}.

\subsection{Symbiotic Generation: Dataset and Avatar}
\label{sec:symgen}
We address the score distillation bottleneck with a straightforward pipeline: reconstruction.
Intuitively, we use video diffusion to generate a portrait dataset with diverse expressions and poses, which serves as pseudo ground-truth for animatable avatar reconstruction.
However, this vanilla implementation often results in poor quality (refer to ``One-time Dataset Update'' in \cref{fig:ablation}).
We attribute this limitation to spatial and temporal inconsistencies introduced by video diffusion in the generated dataset.

To this end, we propose \textbf{Sym}biotic \textbf{GEN}eration, termed as \textbf{SymGEN}, building a mutually beneficial relationship between dataset construction and avatar reconstruction, as illustrated in \cref{fig:pipeline}.
By integrating head avatars into video generation, we enhance dataset quality in spatial and temporal consistency, which in turn refines the detail and fidelity of the avatars trained on this dataset. 
Additionally, we use iterative dataset update~\cite{Ayaan2023instructnerf} that iteratively updates the training dataset and globally consolidates by training the head avatar on the updated dataset.


\noindent\textbf{Avatar-Driven Dataset Enhancement.}
Let $\mathcal{D}^n = \left\{ \left( \mathcal{V}_i, \mathcal{P}_i, \mathcal{E}_i\right) \right\}_{i=1}^{n}$ be the dataset, where $n$ is the number of data.
Each $\mathcal{V}_i$ is a pseudo ground-truth video with corresponding camera sequences $\mathcal{P}_i$ and expression sequences $\mathcal{E}_i$.
Given a sample $\left( \mathcal{V}_i, \mathcal{P}_i, \mathcal{E}_i\right)$, we render a video from the head avatar and extract its Mediapipe facial landmark map~\cite{lugaresi2019mediapipe}.
We then encode the rendered video into a latent code $z_i^0$ using the VAE encoder of the video diffusion, and apply DDIM inversion~\cite{song2020denoising, mokady2023null, liang2024luciddreamer, gaussctrl2024} to obtain its corresponding Gaussian noise $z_i^T$ for texture guidance.
Using the facial landmark map for geometry guidance, we denoise $z_i^T$ to obtain a refined latent code $\hat{z}_i^0$, which is decoded to produce the improved video $\hat{\mathcal{V}}_i$.
Finally, we update the dataset by replacing $\mathcal{V}_i$ with $\hat{\mathcal{V}}_i$.


\noindent\textbf{Dataset-Refined Avatar Reconstruction.}
We then reconstruct the animatable head avatar on the refined synthesized dataset $\mathcal{D}^{n}$.
In specific, we train the avatar with a combination of $\mathcal{L}_1$ and perception loss~\cite{zhang2018perceptual} $\mathcal{L}_\mathrm{LPIPS}$.
Following~\cite{qian2023gaussianavatars}, we incorporate position loss $\mathcal{L}_\mathrm{pos}$ and scaling loss $\mathcal{L}_\mathrm{s}$ to align 3D Gaussians with FLAME, which could effectively avoid outliers.
Formally, the reconstruction loss can be formulated as:
\begin{equation}
    \mathcal{L} = \lambda_\mathrm{1} \mathcal{L}_1 + \lambda_\mathrm{lpips} \mathcal{L}_\mathrm{LPIPS} +  \lambda_\mathrm{pos} \mathcal{L}_\mathrm{pos} + \lambda_\mathrm{s} \mathcal{L}_\mathrm{s},
\label{eq:reconstruction}
\end{equation}
where $\lambda_1 = 10$, $\lambda_\mathrm{lpips} = 10$, $\lambda_\mathrm{pos} = 0.1$ and $\lambda_\mathrm{s} = 10$.

\noindent\textbf{Iterative Mutual Enhancement.}
Following \cite{Ayaan2023instructnerf}, we alternate between avatar-driven dataset enhancement and dataset-refined avatar reconstruction to achieve global consistency. 
Each iteration involves randomly selecting a sample from dataset $\mathcal{D}^{n}$ for avatar reconstruction.
Meanwhile, a dataset update will be performed every $d = 30$ iterations.

\begin{figure*}[t!]
    \centering
    \includegraphics[width=1.0\linewidth]{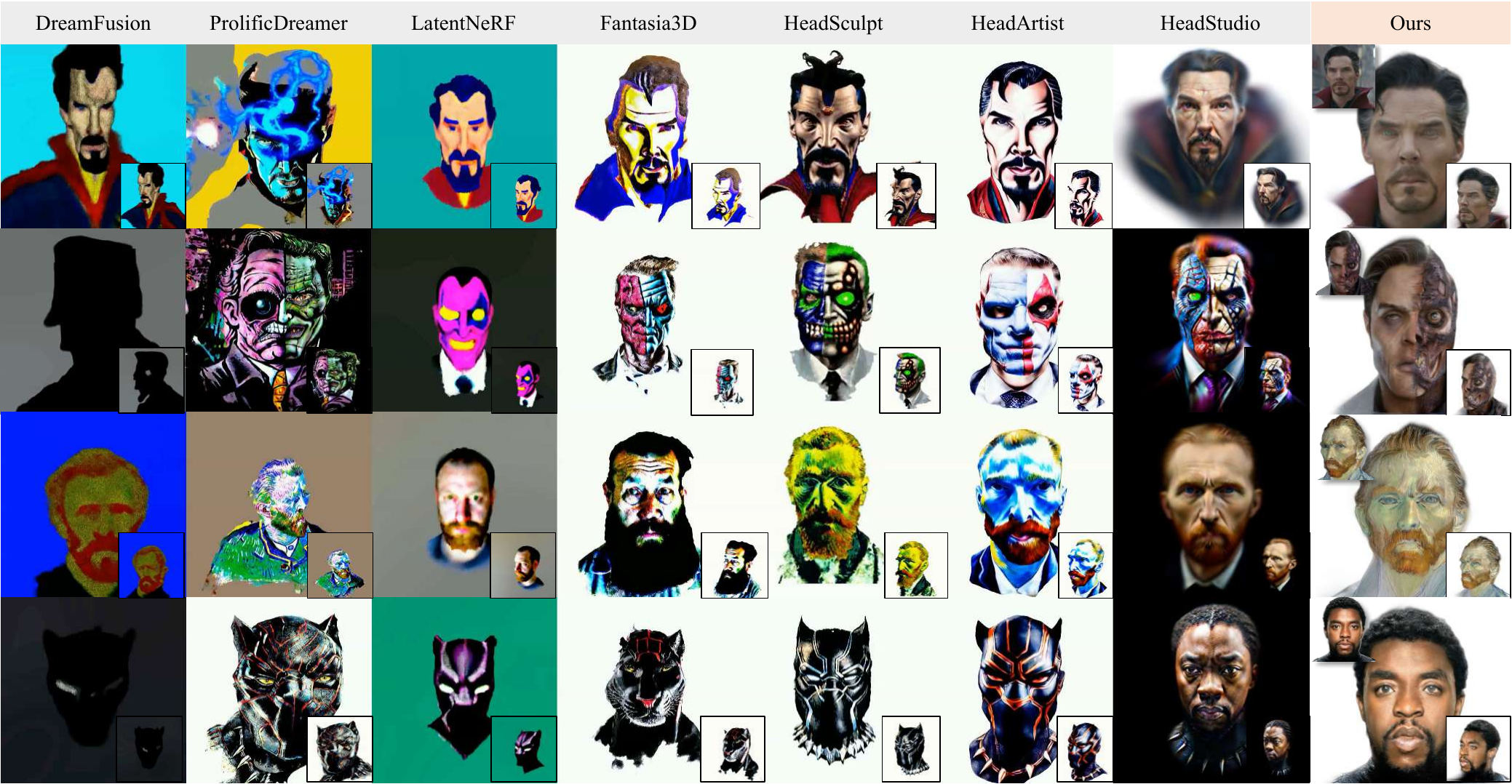}
    \caption{
    \textbf{Comparison with Static Head Avatar Generation Methods.}
    From top to bottom: Doctor Strange, Two-Face from DC, Vincent van Gogh, and Black Panther from Marvel.
    Guided by image prompts, our approach captures rich details and demonstrates superior performance in texture and geometry.
    }
    \vspace{-0.5em}
    \label{fig:static_comp}
\end{figure*}

\subsection{Progressive Learning: Simple to Complex}
\label{sec:prog-learn}
A ``chicken or the egg" dilemma exists in the symbiotic generation.
Initially, low-quality guidance from the avatar results in inconsistent updates to the dataset, which in turn leads to inaccurate reconstructions of the avatar, thus creating a negative cycle.

To break this cycle, we adopt a progressive learning strategy: from simple to complex. 
It is inspired by the observation that \textit{video diffusion yields more consistent results under simple camera poses and relaxed expressions}.
Therefore, we begin with basic camera poses and expressions (\eg, smiling in frontal views) to prevent inconsistencies introduced by video diffusion. 
As the initial head avatar improves, its rendered videos can more effectively guide the video generation under complex expressions and diverse camera angles, thereby enhancing both the dataset and the avatar quality.
Motivated by this insight, we further decouple spatial and temporal in video diffusion into a two stages learning strategy, shown in \cref{fig:progressive}:
1) Spatial Consistency Learning: we fix the expression, progressively learning with camera change from front view to side view.
2) Temporal Consistency Learning: we fix the camera, progressively learning with expression change from relaxed to hyperbole.


\noindent\textbf{Spatial Consistency Learning: Front to Side.}
Let $\mathcal{D}_{S}^{n_s}$ be the spatial dataset of length $n_s$ samples.
For each sample, the fixed expression is associated with an action unit from ARKit blendshape~\cite{ARKit}, representing a basic expression like eye closure and mouth opening.
We then synthesize a camera trajectory $\hat{\mathcal{P}} = \left\{ \hat{p}_i \right\}_{i=1}^{n_f}$ spanning $n_{f}$ video frames.
The trajectory begins with the front view, ends with a randomly sampled side view, and includes interpolations between these viewpoints.

During training, we progressively update the $\mathcal{P} = \left\{ p_i \right\}_{i=1}^{n_f}$ based on the pre-defined camera trajectory $\hat{\mathcal{P}}$ to gradually introduce more complex views as training iterations increase.
Formally, we have:
\begin{equation}
    \begin{split}
        p_i & = \hat{p}_{\min(i, j)}, \\
        j & = \min(\lfloor \frac{k}{d_s} \rfloor + 1, n_f), 
    \end{split}
\end{equation}
where $k$ represents the training iteration and $d_s$ defines the iteration interval.
The variable $j$ ensures that the camera view gradually transitions across the trajectory, initially focusing on the front view and gradually incorporating more challenging views.

\noindent\textbf{Temporal Consistency Learning: Relax to Hyperbole.}
Let $\mathcal{D}_T^{n_t}$ be the temporal dataset with $n_t$ samples.
We use a fixed near-frontal camera pose for each sample in $\mathcal{D}_T^{n_t}$ to focus on temporally expression change.
The temporal dataset begins with simple, relaxed expressions synthesized from~\cite{yi2022generating}, which are easier for the avatar to learn and help mitigate temporal inconsistencies.
As training progresses, we augment the dataset with real-world expressions $\mathcal{E}_\mathrm{real}$, sampling from talk show videos~\cite{yi2022generating}.
Compared to the synthesized expressions, these real-world expression sequences are more exaggerated and challenging, providing greater variability and realism.
This transition is akin to test-time training~\cite{liang2025comprehensive}, allowing the avatar to adapt to more challenging and realistic scenarios, thereby enhancing its robustness and real-world applicability.

\begin{figure*}[t]
    \centering
    \includegraphics[width=1.0\linewidth]{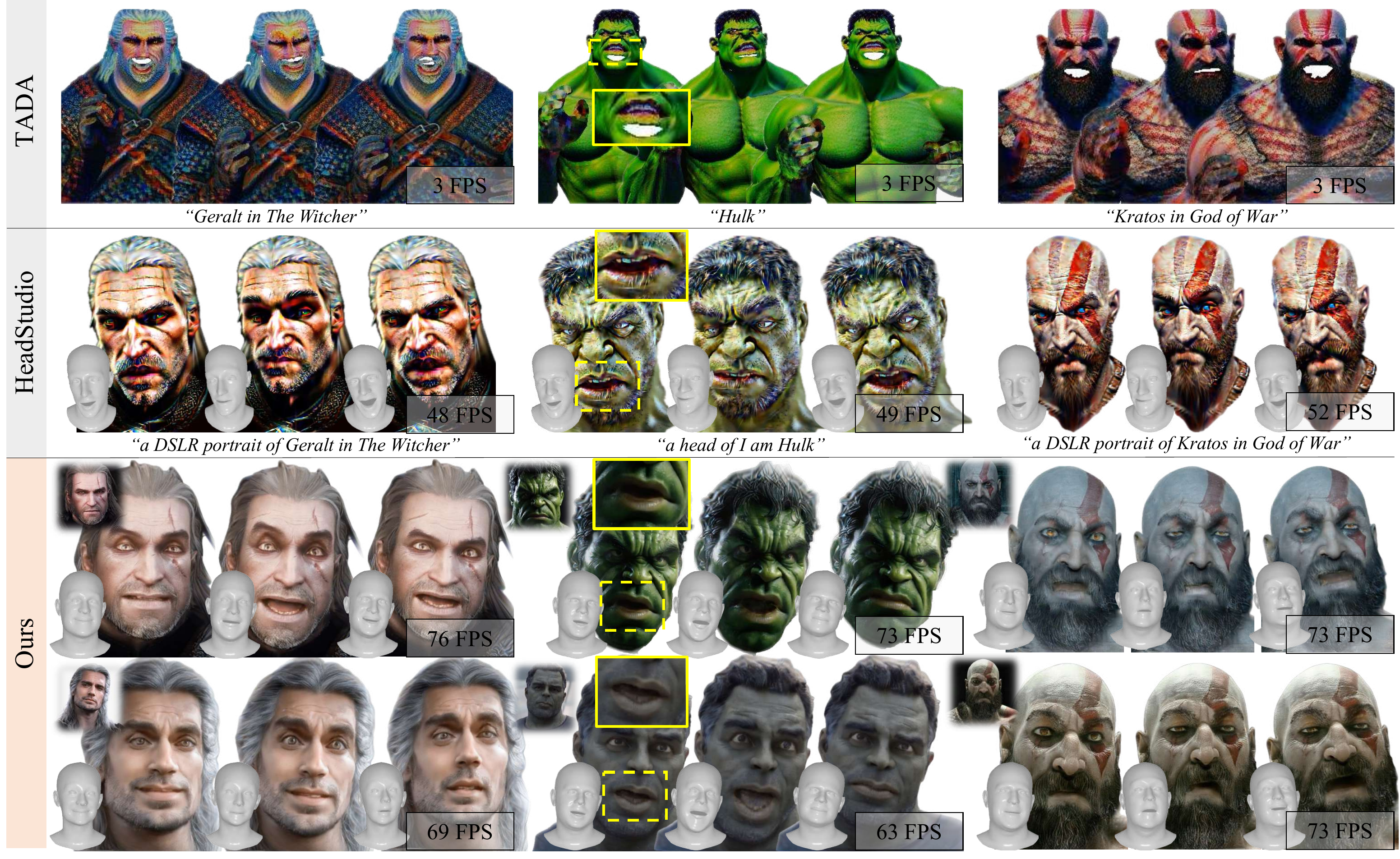}
    \vspace{-1.5em}
    \caption{
    \textbf{Comparison with Dynamic Head Avatar Generation Methods.}
    Yellow circles highlight mouth expression artifacts. 
    Rendering speed on the same device is shown in the black box. 
    The FLAME mesh of the avatar is visualized bottom left. 
    Our method excels in animation quality and rendering speed compared to prior methods.
    }
    \vspace{-0.5em}
    \label{fig:dynamic_comp}
\end{figure*}

\subsection{Implementation Details}
\label{sec:implementation}

\noindent\textbf{Symbiotic Generation.} We use animatable Gaussian head in~\cite{qian2023gaussianavatars} as our 4D avatar. 
Following~\cite{zhou2024headstudio}, we initialize each mesh triangle with 10 evenly distributed  3D Gaussians for faster convergence.
Additionally, we increase the initial opacity of 3D Gaussians near the eyelid, effectively preventing undesired transparency and enabling natural eye closure.
Our method is compatible with portrait video diffusion models driven by facial landmark maps~\cite{wei2024aniportrait, ma2024emoji, chen2024echomimic}.
In our default setup, we use the video diffusion model from~\cite{ma2024emoji} with a default classifier-free guidance (CFG)~\cite{ho2022classifier} weight of $w = 3.5$.

\noindent\textbf{Progressive Learning.}
The entire training consists of $10,000$ iterations.
Initially, only the spatial dataset ($\mathcal{D}^n = \mathcal{D}_S^{n_s}$) with $n_s = 20$ samples and progressively update with $d_s = 1,000$ iterations.
After $k_s = 5,000$ iterations, the temporal dataset with only synthetic expressions is added ($\mathcal{D}^n = \mathcal{D}_S^{n_s} \cup \mathcal{D}_T^{n_\mathrm{syn}}$) with $n_{\mathrm{syn}} = 10$. 
After $k_t = 8,000$ iterations, we expand the temporal dataset by adding $n_\mathrm{real} = 10$ samples of real-world data ($\mathcal{D}^n = \mathcal{D}_S^{n_s} \cup \mathcal{D}_T^{n_\mathrm{syn} + n_\mathrm{real}}$).
We construct the spatial subset by a sampled side view with camera distance range of $\left[8, 11\right]$, a fovy range of $\left[40^\circ, 70^\circ\right]$, an elevation range of $\left[-10^\circ, 10^\circ\right]$, and an azimuth range of $\left[20^\circ, 160^\circ\right]$.
Furthermore, we construct the temporal subset by a sampled fixed camera pose with camera distance range of $\left[8.5, 9.5\right]$, a fovy range of $\left[50^\circ, 60^\circ\right]$, an elevation range of $\left[-10^\circ, 10^\circ\right]$, and an azimuth range of $\left[60^\circ, 120^\circ\right]$.

\noindent\textbf{Training.}
The framework is implemented in PyTorch and threestudio~\cite{threestudio2023}.
We train head avatars with a resolution of $512$ and a batch size of $n_f=8$. 
The entire optimization takes around five hours on a single NVIDIA A6000 (48GB) GPU.
We use the Adam optimizer~\cite{kingma2014adam}, with betas of $\left[0.9, 0.99\right]$, and the learning rates of animatable Gaussian head is set following ~\cite{zhou2024headstudio}.

%% file: sec/4_experiment.tex
\section{Experiment}
\label{sec:exp}
In this section, we first evaluate our method on a range of head avatar generation tasks, then validate our design choices through comparisons with ablated variants, and lastly analyze the limitations of our approach. 
Additional experimental details and results are provided in the supplementary material.

\begin{table}[t]
\begin{center}
\caption{
\textbf{Quantitative Evaluation.}
Evaluating the coherence of generations with their caption using different CLIP models.
}
\vspace{-0.5em}
\begin{tabular}{l|ccc}
\small CLIP-Score & \small ViT-L/14$\uparrow$  & \small ViT-B/16 $\uparrow$ & \small ViT-B/32 $\uparrow$ \\
\hline
\small DreamFusion~\cite{poole2022dreamfusion} & 0.244 & 0.302 & 0.300 \\
\small LatentNeRF~\cite{metzer2022latent} & 0.248 & 0.299 & 0.303 \\
\small Fantasia3D~\cite{Chen2023fantasia3D} & 0.267 & 0.304 & 0.300 \\
\small ProlificDreamer~\cite{wang2023prolificdreamer} & 0.268 & 0.320 & 0.308 \\
\hline
\small HeadSculpt~\cite{han2023headsculpt} & 0.264 & 0.306 & 0.305 \\
\small HeadArtist~\cite{liu2023headartist} & 0.272 & 0.318 & 0.313 \\
\small HeadStudio~\cite{zhou2024headstudio} & 0.275 & \textbf{0.322} & 0.317 \\
\hline
\small Ours & \textbf{0.285} & 0.320 & \textbf{0.322} 
\label{tab:quan}
\end{tabular}
\end{center}
\vspace{-2.0em}
\end{table}

\begin{figure*}
    \centering
    \includegraphics[width=1.0\linewidth]{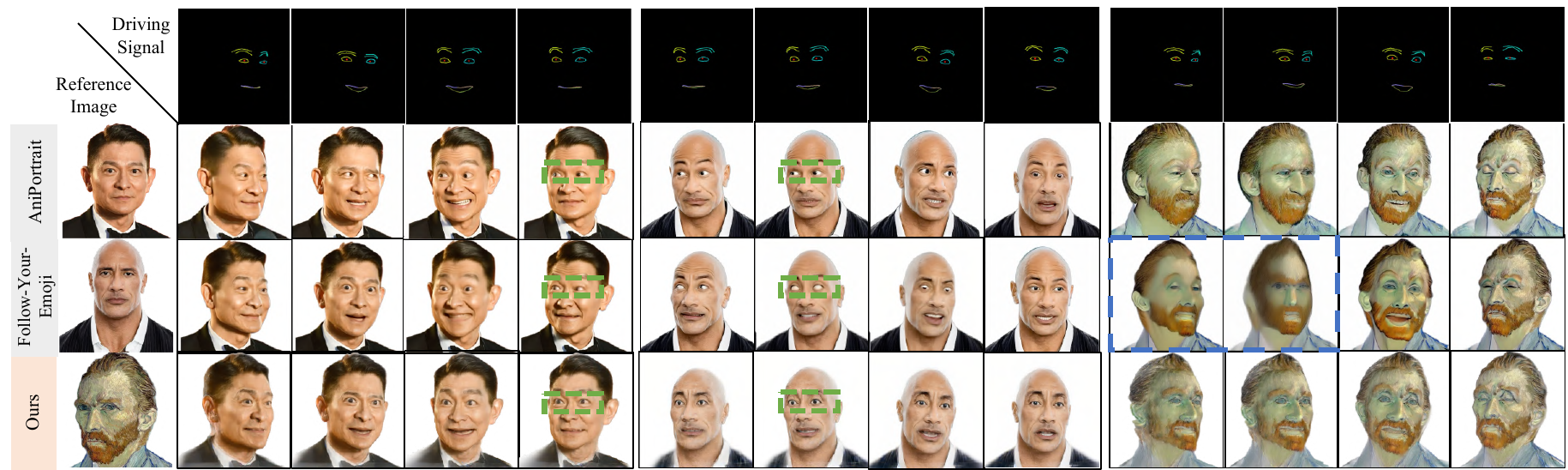}
    \caption{
    \textbf{Comparison with Portrait Video Diffusion Methods.}
    Symbiotic generation enhances portrait video diffusion with improved 3D consistency, temporal smoothness, and expression accuracy. 
    In contrast, traditional portrait video diffusion shows spatial inconsistencies, noted by incorrect eye positioning in side views (green boxes), and temporal inconsistencies, highlighted by significant changes with minor facial expressions (blue boxes).
    }
    \vspace{-1.0em}
    \label{fig:talking_head}
\end{figure*}

\begin{figure*}[t]
    \centering
    \includegraphics[width=1.0\linewidth]{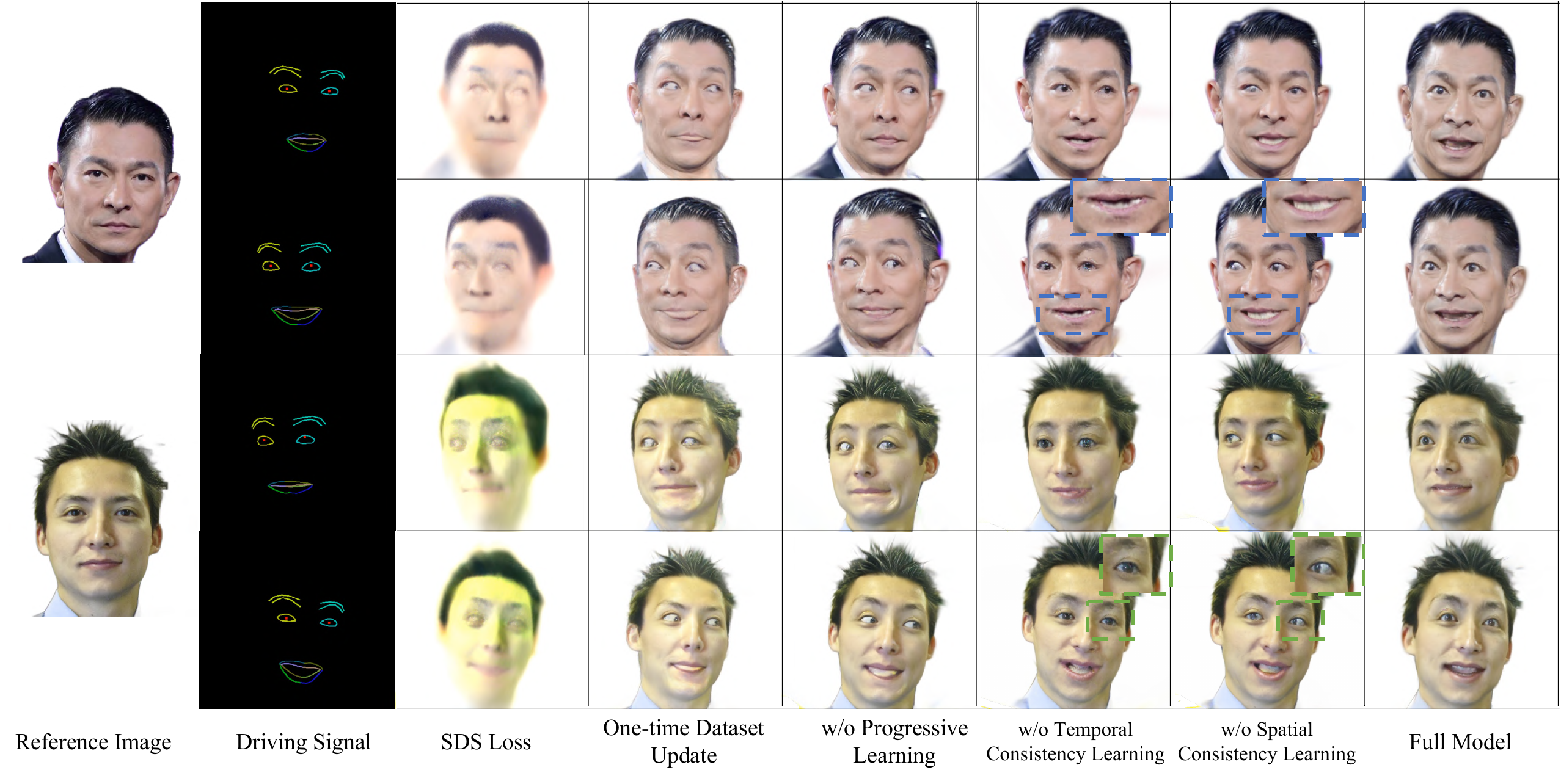}
    \vspace{-2.0em}
    \caption{
    \textbf{Ablation Study.}
    Progressive learning is crucial for creating consistent avatars from inconsistent video diffusion, with spatial consistency improving eye and mouth for effective avatar control (green boxes), and temporal consistency enhancing model generalization to new expressions (blue boxes).
    }
    \vspace{-1.0em}
    \label{fig:ablation}
\end{figure*}

\subsection{Avatar Generation}
As shown in \cref{fig:teaser}, our method is able to generate 4D avatars with fidelity texture and high rendering speed with only one image input.
Meanwhile, it can handle various portrait styles, such as realism, comic, cartoon, \etc.
As a result, we can effectively control the avatar style, such as Joker in realism and comic style.
Following~\cite{zhou2024headstudio}, we evaluate the quality of head avatars with state-of-the-art diffusion-based methods in two settings.

\noindent\textbf{Evaluation of Static Head Avatars.} 
We compare our method with seven avatar generation methods~\cite{poole2022dreamfusion, metzer2022latent, Chen2023fantasia3D, wang2023prolificdreamer, han2023headsculpt, liu2023headartist, zhou2024headstudio}.
Note that HeadSculpt~\cite{han2023headsculpt}, HeadArtist~\cite{liu2023headartist}, and HeadStudio~\cite{zhou2024headstudio} are head-specialized methods. 
As shown in \cref{fig:static_comp}, our method demonstrates superior texture and geometry quality. 
Guided by image prompts, our approach captures rich details, such as the intricate features of Two-Face from DC Comics. 
Following~\cite{liu2023headartist, zhou2024headstudio}, we report the average CLIP score for 10 portraits in \cref{tab:quan}. 
Compared to state-of-the-art methods, our approach shows a $0.1$ and $0.05$ improvement in ViT-L/14 and ViT-B/32 evaluations, respectively, highlighting its effectiveness in high-fidelity avatar generation.

\noindent\textbf{Evaluation of Dynamic Head Avatars.}
We show the comparison with TADA~\cite{liao2023tada} and HeadStudio~\cite{zhou2024headstudio} in \cref{fig:dynamic_comp}.
Our method shows better performance in both animation and rendering speed.
Specifically, compared to HeadStudio~\cite{zhou2024headstudio}, our method reduces artifacts during animation, as highlighted by the blue boxes. 
Additionally, our approach achieves faster rendering speeds (\eg, 76 FPS \emph{vs}\onedot 48 FPS in the Geralt). 
This suggests that symbiotic generation provides more stable and accurate supervision than score distillation-based loss, enabling more precise digital human modeling with fewer Gaussian points.

\subsection{Ablation Study}

\begin{figure*}[t]
    \centering
    \includegraphics[width=1.0\linewidth]{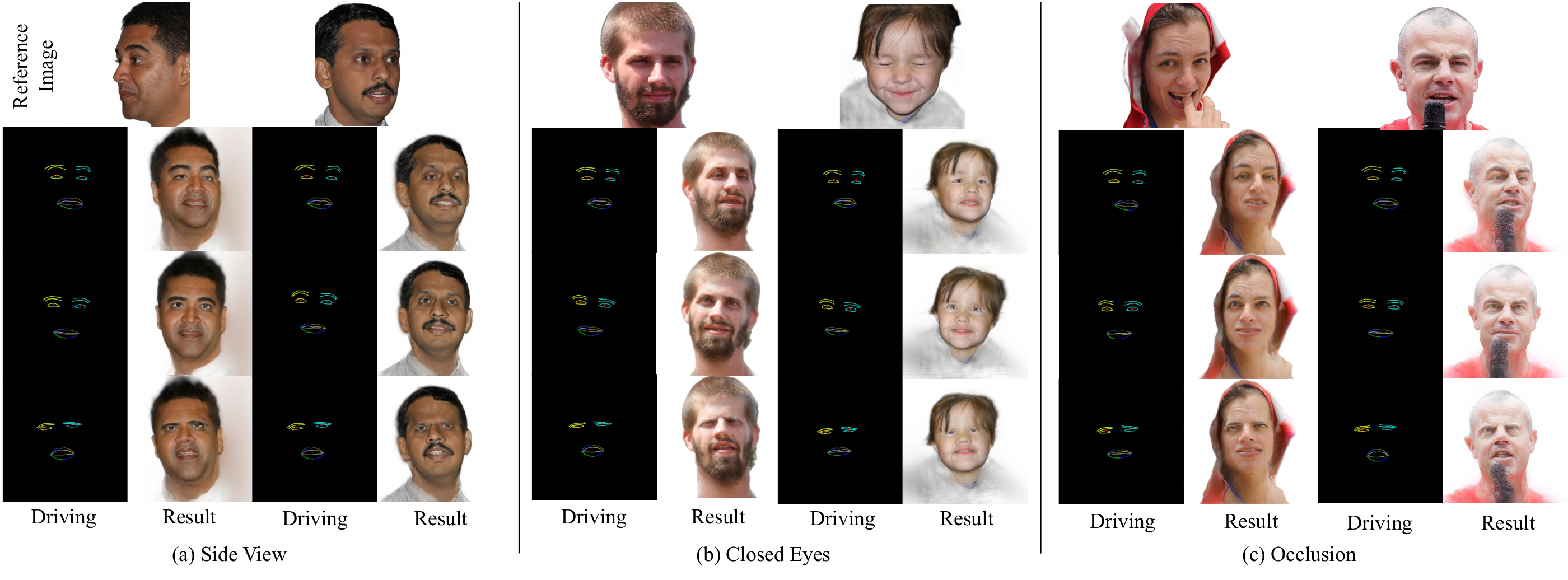}
    \vspace{-2.0em}
    \caption{
    \textbf{Challenge Cases.}
    Our method exhibits robustness in effectively handling side views (\textit{left}), eye closure (\textit{middle}), and facial occlusions (\textit{right}).
    Each pair shows the driving expression and animation result (right), and the top row contains reference images.
    }
    \vspace{-1.0em}
    \label{fig:challenge}
\end{figure*}

\noindent\textbf{Evaluation on Challenge Cases.}
We evaluate Zero-1-to-A on challenging cases, including side views, closed eyes, and facial occlusions. 
As shown in \cref{fig:challenge}, our method demonstrates its effectiveness and robustness by reconstructing full faces from side views, restoring eyes from closed ones, and generating 4D avatars unaffected by occlusions, ensuring seamless animation.

\noindent\textbf{The Effect of Symbiotic Generation.}
To demonstrate the effectiveness of symbiotic generation, we compare it with the portrait video diffusion methods~\cite{wei2024aniportrait, ma2024emoji}, as illustrated in \cref{fig:talking_head}. 
These baseline methods suffer from both spatial and temporal inconsistencies.
As highlighted in the green boxes, there is an incorrect eye positioning in side views, leading to spatial inconsistencies. 
Additionally, as indicated by the blue boxes, minor changes in facial expression can lead to noticeable variations in adjacent frames.
In contrast, our method leverages SymGEN to learn a 4D avatar from these video diffusion methods. 
Compared to the baselines, our approach achieves superior 3D consistency, temporal smoothness, and expression control.

\noindent\textbf{The Effect of Progressive Learning.}
As shown in \cref{fig:ablation}, we evaluate the effectiveness of each component in progressive learning.

\noindent\textit{Video Diffusion + SDS Loss.}
Using only SDS loss~\cite{poole2022dreamfusion} with video diffusion produces over-smooth avatars that lack textural detail.
We attribute this to the spatial and temporal inconsistency in video diffusion, leading to subpar results, which forces us to explore other solutions for diffusion-based generation.

\noindent\textit{One-time Dataset Update.}
In this setup, we use video diffusion to generate images with diverse expressions and poses, serving as pseudo ground-truths for animatable avatar reconstruction. 
The additional dataset helps to alleviate inconsistencies; however, lacking symbiotic generation and progressive learning, the reconstructed avatar shows overly smooth textures and reduced animation quality, particularly in the eyes and mouth.

\noindent\textit{w/o Progressive Learning.} 
In this setup, we utilize the base symbiotic generation with a random learning strategy, where expression and camera sequences are randomly sampled.
While iterative dataset updates~\cite{Ayaan2023instructnerf} improve convergence, the animation quality degrades due to inconsistencies introduced by video diffusion. 
Notably, misalignments in the eyes and teeth of the avatar result in inaccurate expression control.
These results highlight the importance of progressive learning in constructing a consistent avatar from inconsistent video diffusion outputs.

\noindent\textit{w/o Temporal Consistency Learning.}
In this setup, only spatial consistency learning is applied. 
Experimental results show improved eye and mouth alignment.
With a progressive strategy, starting from frontal views to side views, the model learns reasonably accurate eye and mouth, allowing for effective avatar control.
However, without temporal consistency learning, the avatar lacks generalizability for animation. 
For hyperbole expressions (highlighted in the blue boxes), significant artifacts appear in mouth deformation due to the absence of temporal smoothness.

\noindent\textit{w/o Spatial Consistency Learning.}: 
Applying only temporal consistency learning improves generalization and reduces artifacts in hyperbole expressions.
However, misaligned eyes (highlighted in the green boxes) indicate the necessity of spatial consistency learning.
These findings also suggest that incorporating challenging samples (\eg, large angles and hyperbole expressions) into temporal consistency learning, akin to test-time training~\cite{sun2020test}, could further enhance generalization under distribution shifts.

\begin{figure}
    \centering
    \includegraphics[width=1.0\linewidth]{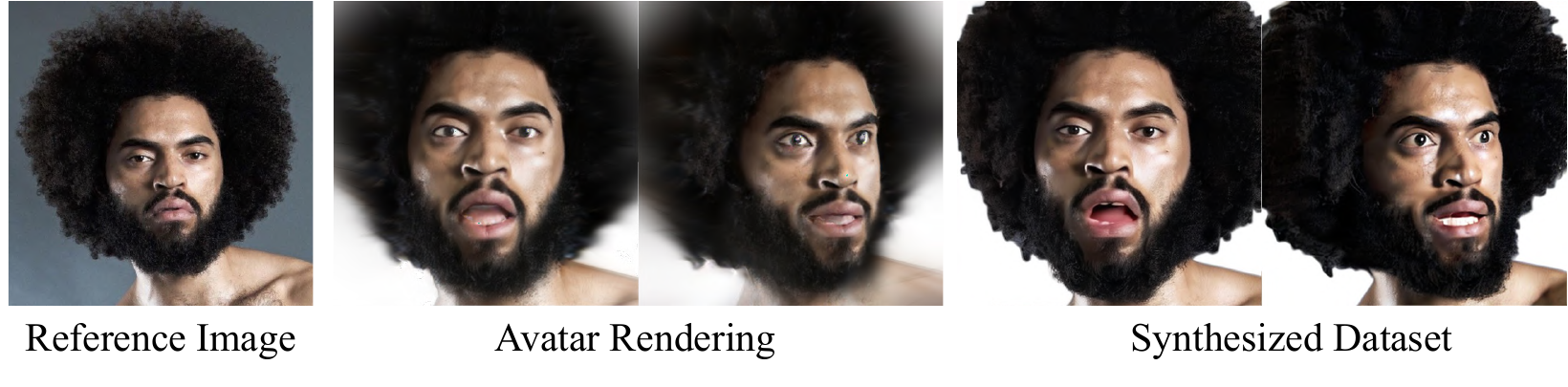}
    \vspace{-1.5em}
    \caption{
    \textbf{Limitation.} The animatable Gaussian head~\cite{qian2023gaussianavatars} aligns Gaussians with the FLAME mesh, limiting the modeling of elements beyond the head.
    }
    \vspace{-1.0em}
    \label{fig:limitation}
\end{figure}

\subsection{Limitations}
Our method excels in texture and geometry, and maintains improved spatial-temporal consistency in facial animation.
However, it encounters challenges in effectively modeling elements beyond the head.
As shown in \cref{fig:limitation}, it inadequately models a man with an afro hairstyle.
This limitation arises from the animatable Gaussian head~\cite{qian2023gaussianavatars,zhou2024headstudio}, which constrains Gaussians to align with the FLAME mesh, leading to the blurred afro.
Additionally, edge blur may result from underfitting and labeling ambiguities.
To mitigate these issues, we recommend additional rapid reconstruction on key frames and incorporating complementary representations such as Gaussian hair~\cite{luo2024gaussianhair}.

%% file: sec/5_conclusion.tex
\section{Conclusion}
\label{sec:conclusion}


In this paper, we proposed Zero-1-to-A, a novel method for generating high-fidelity 4D avatars from a single image using pre-trained video diffusion models. Zero-1-to-A tackles spatial and temporal inconsistencies by iteratively synthesizing consistent datasets and employing a Progressive Learning strategy, which ensures stable initialization and smooth quality improvement. Experiments show that Zero-1-to-A outperforms existing methods in fidelity, animation quality, and rendering speed, providing a robust and data-efficient solution for lifelike avatar creation.

%% file: sec/X_suppl.tex
\clearpage
\setcounter{page}{1}
\maketitlesupplementary


\section{Introduction}
In this supplementary material, we provide additional details and insights into the work presented in the paper.

\begin{itemize} 
\item \cref{sec:supp-imply} details the image preprocessing steps and explores potential applications for generating avatars.
\item \cref{sec:discussion} discusses related solutions, highlighting their limitations, the differences from our approach, and the advantages these differences bring.
\item \cref{sec:supp-exp} presents comprehensive experimental results, including qualitative comparisons with image-to-3D methods, quantitative evaluations against video diffusion models, and various ablation studies.
\end{itemize}

Furthermore, we visualize the spatial and temporal inconsistencies in video diffusion models and demonstrate the improvements introduced by our method (\cref{sec:dis_motivation}). In the ablation study, we evaluate the adaptability of our method to different video diffusion models and its applicability to in-the-wild datasets.

\section{Additional Implementation Details}
\label{sec:supp-imply}

\begin{figure}[t]
    \centering
    \includegraphics[width=1.0\linewidth]{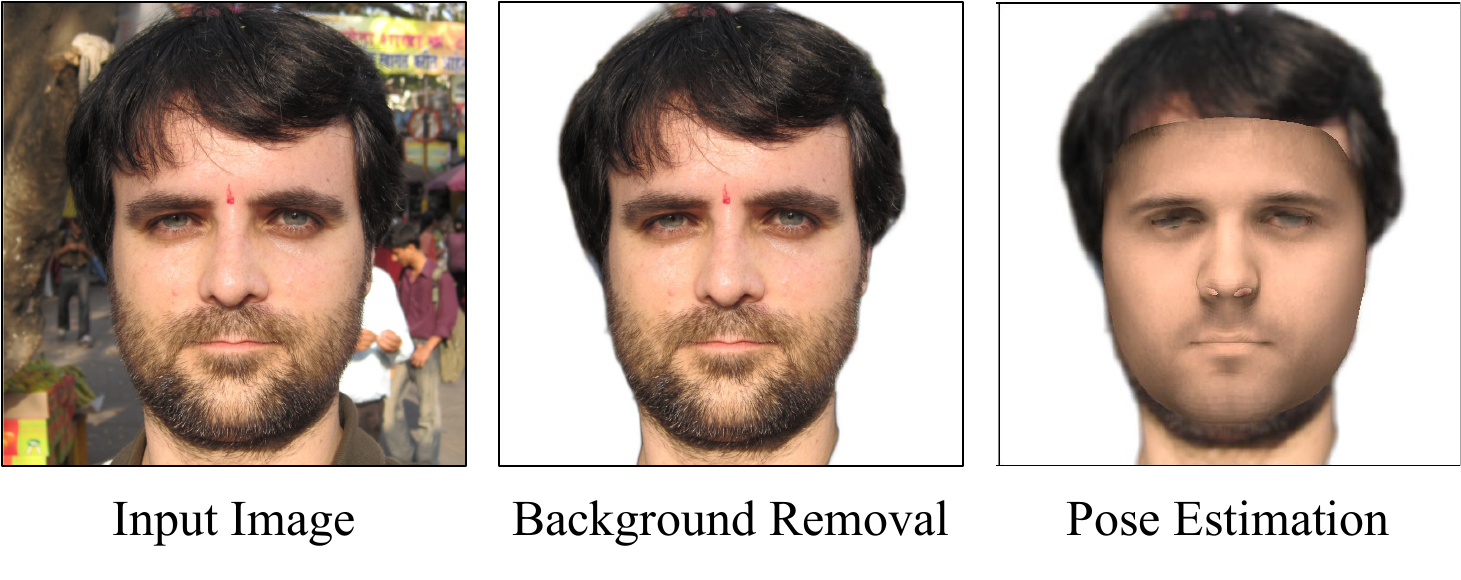}
    \caption{
    \textbf{Pre-process.}
    Given a portrait image, we first remove its background and then estimate the FLAME parameter.
    }
    \label{fig:preprocess}
\end{figure}

\noindent\textbf{Pre-process.}
Based on \cite{qin2020u2, chu2024gpavatar, EMOCA:CVPR:2021}, the preprocessor removes the background and estimates the pose and FLAME parameters of input portrait.
As shown in ~\cref{fig:preprocess}, the pre-process of each image contains two steps:
1) Background Removal: Given a portrait image, we first use Rembg~\cite{qin2020u2} to remove the background and only retain the foreground portrait;
2) Pose Estimation: Then, we use MICA~\cite{MICA:ECCV2022} and EMOCA~\cite{EMOCA:CVPR:2021} to estimate the FLAME shape, expression and pose parameters of portrait.
In training, we use the carved image as the input of video diffusion and initialize the learnable shape parameter with estimated FLAME shape.

\noindent\textbf{Application.}
Once optimized, the parameters of the animatable Gaussian head are fixed, enabling real-time animation and rendering of the avatar using motion and camera sequences. 
These motion sequences incorporate expression and pose parameters from FLAME 2020~\cite{li2017flame}. 
Following HeadStudio~\cite{zhou2024headstudio}, we employ advanced models such as face-to-FLAME~\cite{DECA:Siggraph2021, EMOCA:CVPR:2021, MICA:ECCV2022}, speech-to-FLAME~\cite{yi2022generating}, and text-to-speech~\cite{PlayHT} to convert video, speech, and text into FLAME animation inputs.
As a result, the generated avatar supports multimodal control, demonstrating practical applications in real-world scenarios.

\section{Discussion}
\label{sec:discussion}

\subsection{Discussion with Related Solutions}

\begin{figure}[t]
    \centering
    \includegraphics[width=1.0\linewidth]{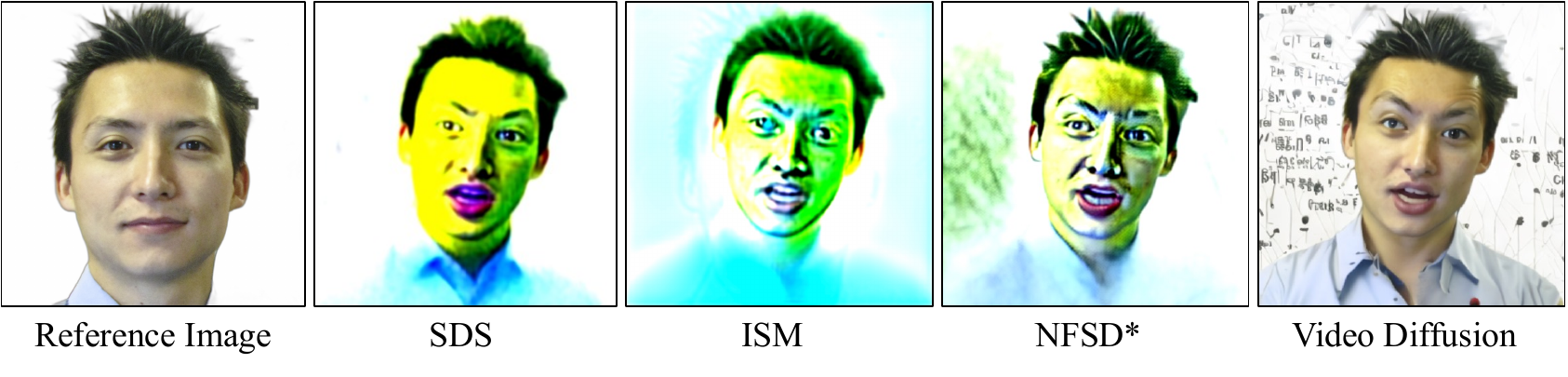}
    \caption{
    \textbf{2D Image Generation with SDS-based Loss.}
    From left to right: reference image, SDS~\cite{poole2022dreamfusion}, ISM~\cite{liang2024luciddreamer}, NFSD~\cite{katzir2023noise} and video diffusion generation~\cite{wei2024aniportrait}.
    }
    \label{fig:comp_sds}
\end{figure}

\begin{figure*}[t]
    \centering
    \includegraphics[width=1.0\linewidth]{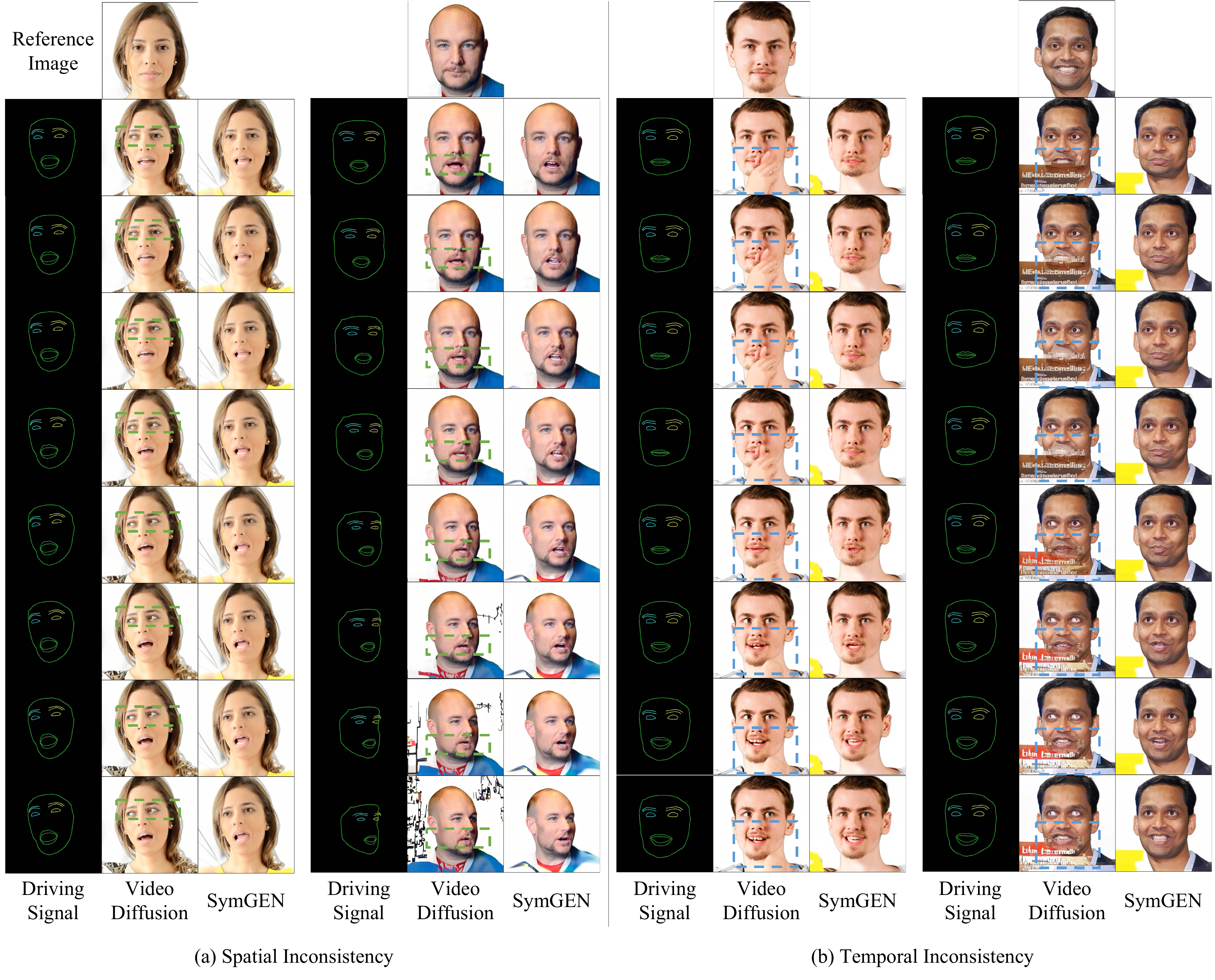}
    \vspace{-2.0em}
    \caption{
    \textbf{Visualization of Spatial and Temporal Inconsistencies in Video Diffusion Models.}
    Portrait video diffusion exhibits spatial inconsistencies, such as incorrect eye positioning in side views (green boxes), and temporal inconsistencies, evident in significant changes triggered by minor facial expressions (blue boxes).
    }
    \vspace{-1.0em}
    \label{fig:motivation}
\end{figure*}

\noindent\textbf{DreamFusion v.s. Zero-1-to-A.}
In \cref{fig:comp_sds}, we compare 2D image generation results using SDS~\cite{poole2022dreamfusion}, ISM~\cite{liang2024luciddreamer}, and NFSD~\cite{katzir2023noise}. 
Notably, NFSD employs negative prompts to suppress unwanted noise in the diffusion score. We implement this by treating reference image with data augmentation (e.g., blur, brightness adjustment, Gaussian noise) as negative prompts.
Compared to video diffusion generation, results from SDS-based loss exhibit issues of over-smoothing and over-saturation. 
We attribute this to additional temporal modules introduced in portrait video diffusion, which may adversely affect score distillation.
This limitation motivates us to explore alternative solutions.

\noindent\textbf{Instruct-NeRF2NeRF (IN2N) v.s. Zero-1-to-A.}
IN2N~\cite{Ayaan2023instructnerf} introduces a 3D editing method called iterative dataset update, which alternates between editing the ground-truth dataset and optimizing the 3D scene.
In contrast, Zero-1-to-A is a 4D generation method that progressively builds a pseudo ground-truth dataset while optimizing the 4D avatar.
Different from the editing task, the lack of consistent input in the generation task creates a negative cycle in iterative dataset update, leading to incorrect convergence (\eg, misaligned eyes and inability to open the mouth, as shown in the fifth column of Fig.~\textcolor{blue}{8}).
To address this, we propose a simple-to-complex progressive learning strategy that breaks this cycle and significantly improves generation performance.

\begin{figure*}
    \centering
    \includegraphics[width=1.0\linewidth]{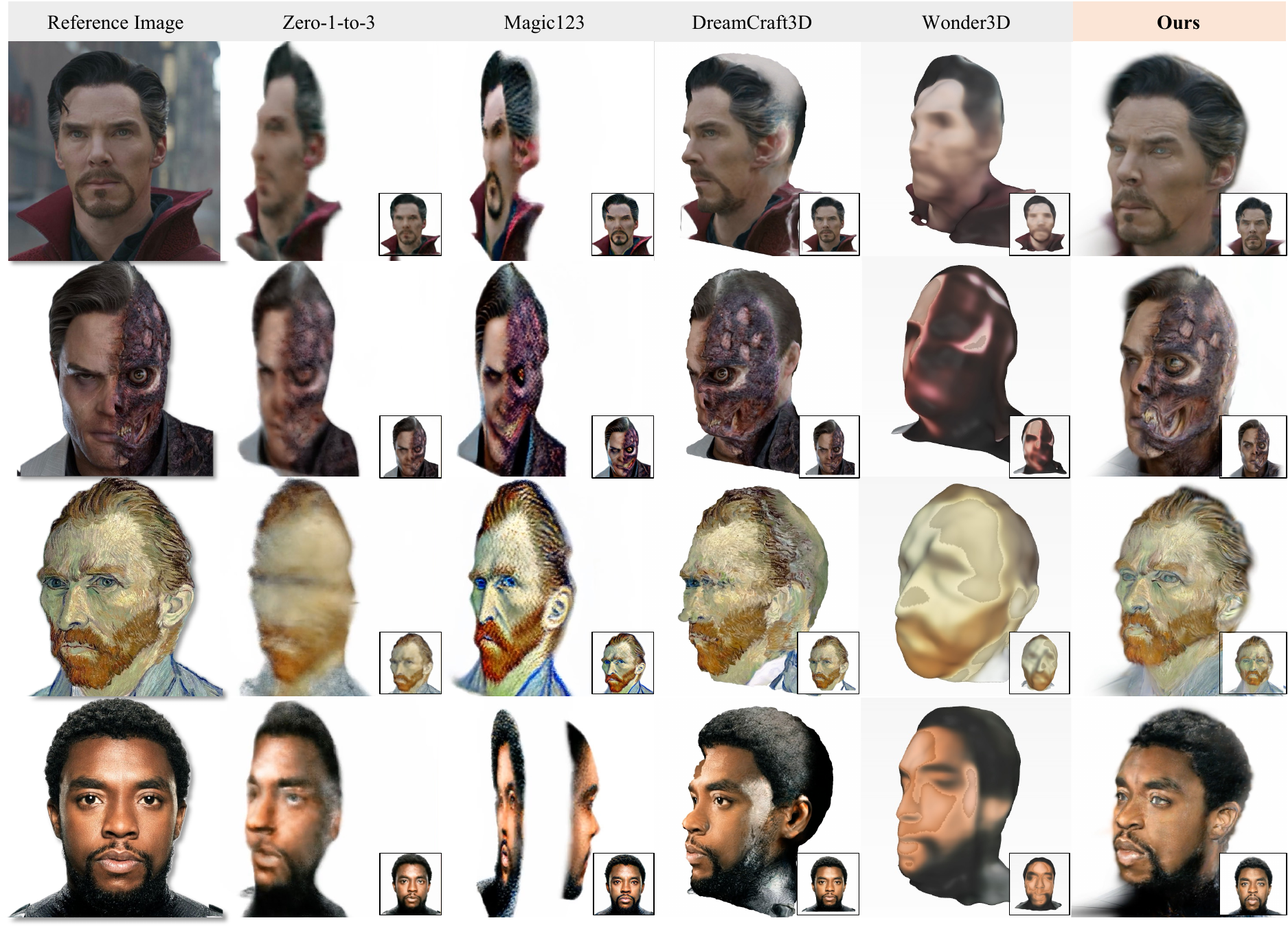}
    \caption{
    \textbf{Comparisons with Image-to-3D Methods.}
    Our method delivers comparable performance in texture reconstruction while achieving superior 3D consistency.
    }
    \vspace{-1.0em}
    \label{fig:comp_static_image}
\end{figure*}
\begin{figure}[t]
    \centering
    \includegraphics[width=1.0\linewidth]{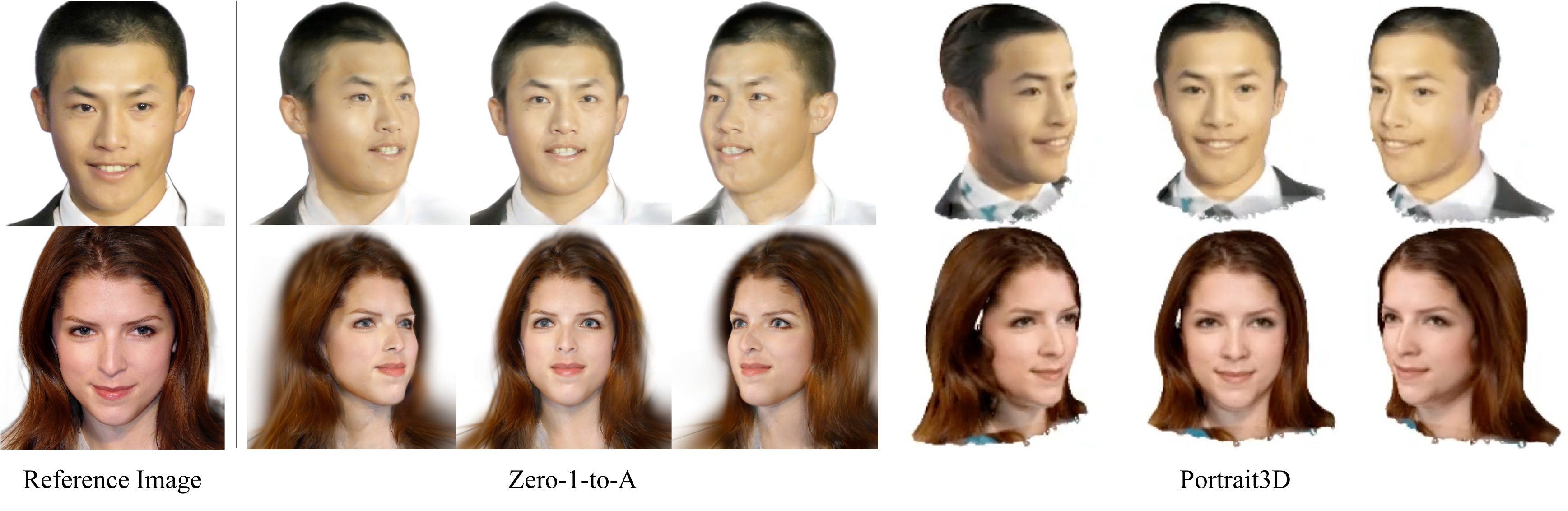}
    \vspace{-2.0em}
    \caption{
    \textbf{Comparisons with Portrait3D~\cite{hao2024portrait3d}.}
    Our method matches the performance of Portrait3D while providing animatable avatars, enabling a wider range of applications.
    }
    \vspace{-1.0em}
    \label{fig:comp_portrait3d}
\end{figure}

\begin{figure}[t]
    \centering
    \includegraphics[width=1.0\linewidth]{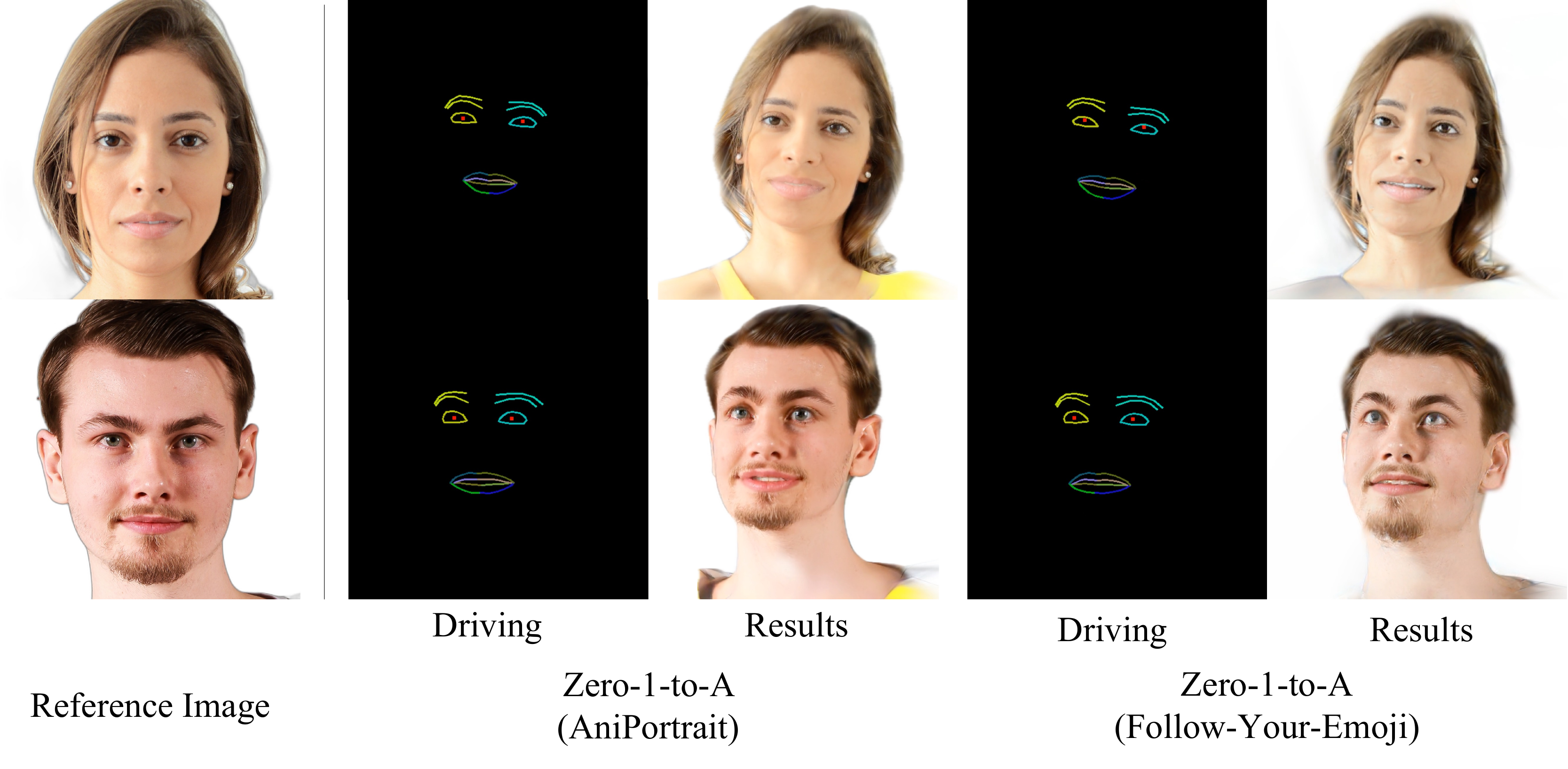}
    \vspace{-2.0em}
    \caption{
    \textbf{Evaluation on Different Video Diffusion Models.}
    Our method demonstrates its effectiveness by seamlessly adapting to various video diffusion models.
    }
    \vspace{-1.0em}
    \label{fig:abl_vid_diff}
\end{figure}

\begin{table}[t]
    \centering
    \caption{
    \textbf{Quantitative Evaluation of Avatar Animation.}
    We evaluate ID consistency, temporal smoothness, and rendering speed, demonstrating that our method is able to enhance the performance of portrait video diffusion.
    }
    \begin{tabular}{c|ccc}
    \hline
    \small Face Animation & \small ID $\uparrow$ & \small Motion $\uparrow$ & \small Speed $\uparrow$ \\
    \hline
    \small AniPortrait~\cite{wei2024aniportrait} & \small \textbf{0.5081} & \small 0.8410 & \small{0.52 FPS} \\
    \small Follow-Your-Emoji~\cite{ma2024emoji} & \small 0.4988 & \small 0.8934 & \small {0.56 FPS} \\
    \hline
    \small Ours (w.~\cite{ma2024emoji}) & \small 0.5000 & \small \textbf{0.9187} & \small{\textbf{71 FPS}} \\
    \hline
    \end{tabular}
    \vspace{-1.0em}
    \label{tab:eval_vid}
\end{table}

\subsection{Discussion on the Motivation}
\label{sec:dis_motivation}


In \cref{fig:motivation}, we show the spatial and temporal inconsistencies in video diffusion models and demonstrate the improvements achieved by our method.
On the left of \cref{fig:motivation}, spatial inconsistencies are shown by fixing the expression and varying the camera pose. 
Ideally, the portrait's expression should remain unchanged. 
However, as the camera pose shifts, the iris incorrectly looks left, and teeth that were initially absent appear (highlighted in green boxes).
On the right, temporal inconsistencies are illustrated by fixing the camera pose and varying the expression. 
Ideally, the portrait should deform smoothly and accurately. 
Instead, even with minor changes, such as gradually opening the mouth, the generated video exhibits abrupt and incorrect variations (highlighted in blue boxes).
With SymGEN, we achieve improvements in video generation under large pose changes and exaggerated expressions, resulting in a spatially and temporally consistent pseudo-ground truth dataset.

In summary, video diffusion models~\cite{wei2024aniportrait, ma2024emoji} suffer from severe spatial and temporal inconsistencies, making them unsuitable for direct 4D avatar reconstruction.
Our proposed SymGEN framework iteratively constructs a consistent dataset, enabling the reconstruction of 4D avatars.

\section{Additional Experiments}
\label{sec:supp-exp}

\subsection{Comparisons with Image-to-3D Methods}
In \cref{fig:comp_static_image}, we compare our method with diffusion-based image-to-3D approaches, including Zero-1-to-3~\cite{liu2023zero}, Magic123~\cite{qian2023magic123}, DreamCraft3D~\cite{sun2023dreamcraft3d}, and Wonder3D~\cite{long2024wonder3d}. 
We reproduce Zero-1-to-3, Magic123, and DreamCraft3D using threestudio\footnote{https://github.com/threestudio-project/threestudio} and implement Wonder3D with NeuS following the official guidelines\footnote{https://github.com/xxlong0/Wonder3D}. 
The results show that our Zero-1-to-A delivers comparable texture fidelity and superior geometry reconstruction, leveraging a head prior model. 

\noindent\textbf{Comparisons with Portrait3D.}
Portrait3D~\cite{hao2024portrait3d} is a diffusion-based image-to-avatar method. 
However, as the code is not yet open-sourced, we could not reproduce its results for the avatar generation benchmark~\cite{liu2023headartist, zhou2024headstudio}. 
In \cref{fig:comp_portrait3d}, we compare our method with Portrait3D using results captured from its official project\footnote{https://jinkun-hao.github.io/Portrait3D/}. 
Our method achieves comparable performance while offering animatable avatars, enabling broader applications than Portrait3D.

\subsection{Comparisons with Video Diffusion Methods.}
In \cref{tab:eval_vid}, we quantitatively evaluate ID consistency, temporal smoothness, and rendering speed. 
ID consistency (ID) is measured using cosine similarity of identity embeddings, while temporal smoothness (Motion) is evaluated using a stability score based on frequency analysis of estimated 2D motion. Higher low-frequency energy indicates greater video stability (details in \cite{liu2013bundled}). 
The evaluation is conducted on 18 samples and 300 frames from real-world portrait videos~\cite{yi2022generating}. 
Our method demonstrates improved ID consistency, temporal smoothness, and rendering speed, highlighting the effectiveness of Zero-1-to-A.

\subsection{Additional Ablations}
\label{sec:supp-dis}

\noindent\textbf{Evaluation on Different Video Diffusion Models.}
In \cref{fig:abl_vid_diff}, we compare results using different video diffusion models (AniPortrait~\cite{wei2024aniportrait} and Follow-Your-Emoji~\cite{ma2024emoji}). 
Notably, using AniPortrait achieves better color fidelity to the reference image than using Follow-Your-Emoji.
Our method adapts seamlessly to various video diffusion models, effectively generating animatable avatars and demonstrating robust performance.

\begin{figure*}[t!]
    \centering
    \includegraphics[width=0.88\linewidth]{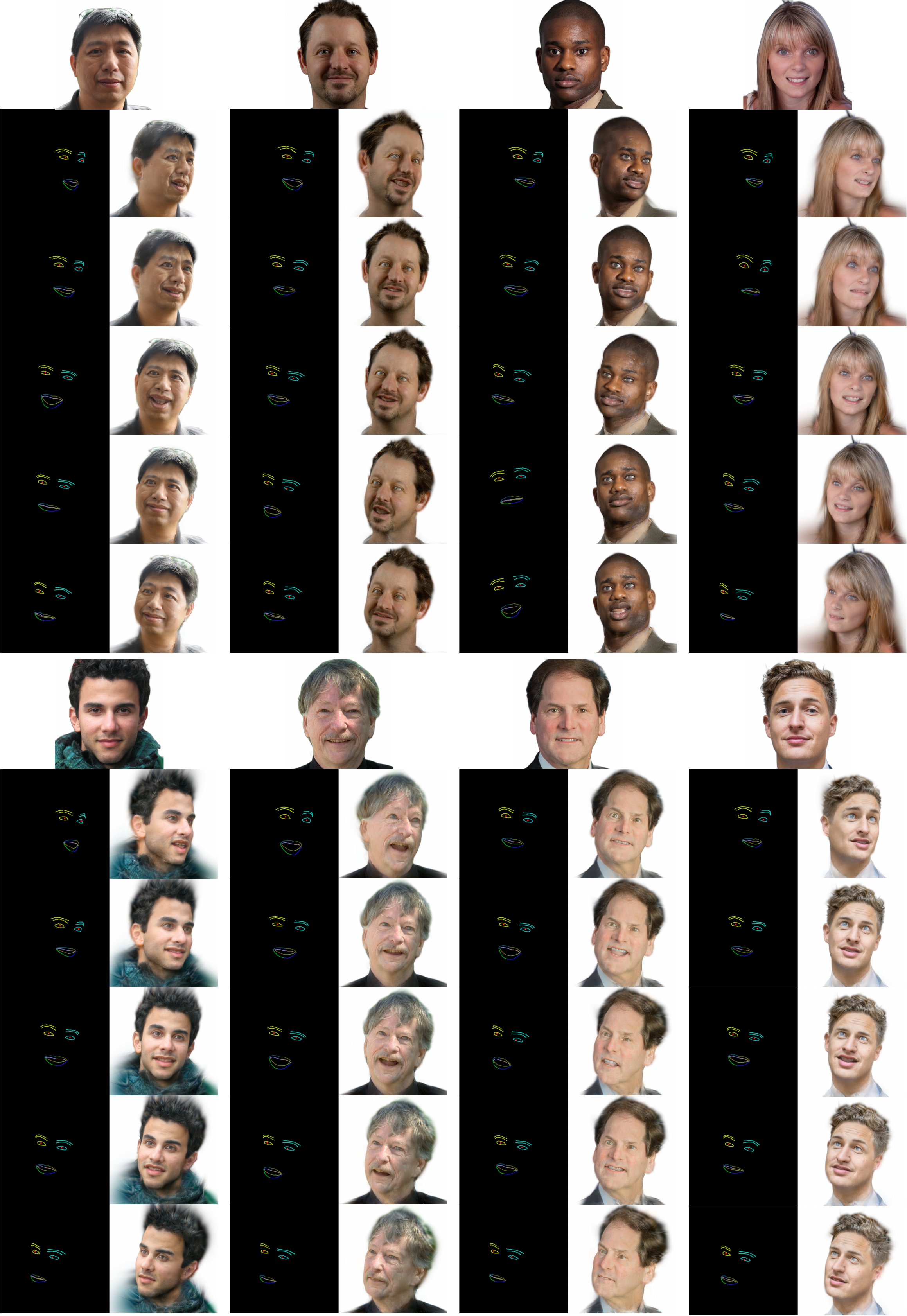}
    \caption{
    \textbf{Evaluation on in-the-wild Cases.}
    }
    \label{fig:in_the_wild}
\end{figure*}

\noindent\textbf{Evaluation on in-the-wild Cases.}
We further evaluate our method on a wide range of in-the-wild cases. 
Specifically, we sampled multiple portraits from the FFHQ dataset~\cite{karras2019style}, with results shown in \cref{fig:in_the_wild}. 
The results demonstrate that our approach is broadly applicable across genders and diverse ethnic groups.

